\documentclass{llncs}

\usepackage{llncsdoc}
\usepackage{graphicx}
\usepackage{color}
\usepackage[dvipsnames]{pstricks}

\newcommand{\keywords}[1]{\par\addvspace\baselineskip
\noindent\keywordname\enspace\ignorespaces#1}

\begin {document}
\title{Solving Rubik's Cube Using SAT Solvers}
\titlerunning{Phase Selection Heuristics for SAT Solvers}

\author{Jingchao Chen}
\institute{School of Informatics, Donghua University \\
2999 North Renmin Road, Songjiang District, Shanghai 201620, P. R.
China \email{chen-jc@dhu.edu.cn}}

\maketitle
\begin{abstract}
Rubik's Cube is an easily-understood puzzle, which is originally
called the ``magic cube". It is a well-known planning problem, which
has been studied for a long time. Yet many simple properties remain
unknown. This paper studies whether modern SAT solvers are
applicable to this puzzle. To our best knowledge, we are the first
to translate Rubik's Cube to a SAT problem. To reduce the number of
variables and clauses needed for the encoding, we replace a naive
approach of 6 Boolean variables to represent each color on each
facelet with a new approach of 3 or 2 Boolean variables. In order to
be able to solve quickly Rubik's Cube, we replace the direct
encoding of 18 turns with the layer encoding of 18-subtype turns
based on 6-type turns. To speed up the solving further, we encode
some properties of two-phase algorithm as an additional constraint,
and restrict some move sequences by adding some constraint clauses.
Using only efficient encoding cannot solve this puzzle. For this
reason, we improve the existing SAT solvers, and develop a new SAT
solver based on PrecoSAT, though it is suited only for Rubik's Cube.
The new SAT solver replaces the lookahead solving strategy with an
ALO (\emph{at-least-one}) solving strategy, and decomposes the
original problem into sub-problems. Each sub-problem is solved by
PrecoSAT. The empirical results demonstrate both our SAT translation
and new solving technique are efficient. Without the efficient SAT
encoding and the new solving technique, Rubik's Cube will not be
able to be solved still by any SAT solver. Using the improved SAT
solver, we can find always a solution of length 20 in a reasonable
time. Although our solver is slower than Kociemba's algorithm using
lookup tables, but does not require a huge lookup table.

\end{abstract}

\section{Introduction}
SAT solvers have attained success in many fields, and have been used
widely for hardware design and verification, software verification,
artificial intelligence, cryptanalysis, equivalence checking, model
checking, planning, scheduling etc. However, there are still large
real world instances that cannot be solved by SAT solvers. Rubik's
Cube is such an example. It is a well-known planning problem, which
is originally called the ``magic cube''. The puzzle game was
invented in 1974[1] by Ern\"{o} Rubik. So far, It has been studied
for a long time. Yet many simple properties remain unknown. From the
viewpoint of SAT applications, this paper studies this puzzle.

 With respect to this puzzle, one of the most natural
questions is how many moves are required to solve Rubik's Cube in
the worst case. This problem has been studied for over 30 years.
There has been great progress. In 1995, Reid proved that the lower
bound on the number of moves and the upper bound is 20 and 29,
respectively \cite{Rokicki:5,Reid:6,Reid:7}. Since then, the upper
bound was unceasingly improved. In 2006, the upper bound was reduced
by Radu \cite{Radu:8} to 27. In 2007, Kunkle and Cooperman used
computer search methods to refine it to 26. In 2008, Rokicki
\cite{Rokicki:3,Rokicki:4} reduced further it from 25 to 22. In
2010, this open problem was settled. Rokicki, Kociemba, Davidson and
Dethridge \cite{Flatley:9,Rokicki:10} proved that God's number (i.e.
the upper bound) for the Cube is exactly 20. They spent about 35
CPU-years of idle computer donated by Google to solve all
43,252,003,274,489,856,000 positions of the Cube. Without a doubt,
all the current approaches to proving the upper bound are
time-consuming and space-consuming. How to finish the theoretical
proof of the upper bound is yet a hard problem.

Apart from the approach to compute directly positions of the Cube,
one considered the other approaches. In 1985, Korf \cite{Korf:11}
noted that problems such as Rubik's Cube can be divided into
subgoals that are of the property called operator decomposability,
and attempted to solve them by searching for macro-operators. Korf's
approach succeeded for the $2\times2\times2$ version of Rubik's
Cube, but failed to find an optimal solution for the full
$3\times3\times3$ Rubik's Cube, for which the solution lengths are
close to those of human strategies. Rubik's Cube is also a
well-known planning problem. This puzzle can be encoded as a
planning problem in PDDL (Planning Domain Definition Language).
Nevertheless, there is no report on solving successfully it using a
sat-based planner such as SATPLAN \cite{Kautz:12}.

The purpose of this paper is two-fold. The first purpose is to
design an effective SAT encoding of Rubik's Cube. The second purpose
is to improve the existing SAT solvers to extend the range of SAT
applications. To attain the first purpose, we optimize the encoding
of this puzzle in the following ways: encoding the At-Most-One (AMO)
constraint for minimizing the number of moves by the 2-product
encoding \cite{chenAMO:14} proposed recently, and replacing a naive
approach of 6 Boolean variables to represent each color on each
facelet with a new approach of 2 Boolean variables. The number of
Boolean variables required by a cube state is cut from
$8\times6\times6=288$ to $8\times6\times3=144$ or
$8\times6\times2=96$. Encoding this puzzle according to the idea of
one-phase algorithm results in a very hard SAT problem. Therefore,
we encode this problem according to the idea of two-phase algorithm,
and consider the goal state of phase 1 as an additional constraint.
In addition to efficiently encoding this puzzle, we improve the
existing SAT solvers, and develop a new hybrid SAT solver based on
PrecoSAT \cite{Precosat:13}, though the new solver is suited only
for Rubik's Cube. The new solver selects some decision variables
according to whether a variable occurs in a clause with the AMO
constraint, and splits the original problem into some subproblems
with those decision variables. Each subproblems is solved by
PrecoSAT. The empirical results demonstrate both our SAT translation
and new solving technique are efficient. Without the efficient SAT
encoding and the new solving technique, Rubik's Cube will not be
able to be solved still by any SAT solver. Using the improved SAT
solver, we can find always a solution of length 20 in a reasonable
time. Although our solving speed is slower than the non-SAT solver
such as Kociemba's algorithm using lookup tables, but does not
require a huge lookup table.  needs less memory. 10 MB RAM memory is
sufficient for the approach to use the SAT solver.

%\nopagebreak[2]
%\samepage
%\linepenalty=-1000
\section{Preliminaries}

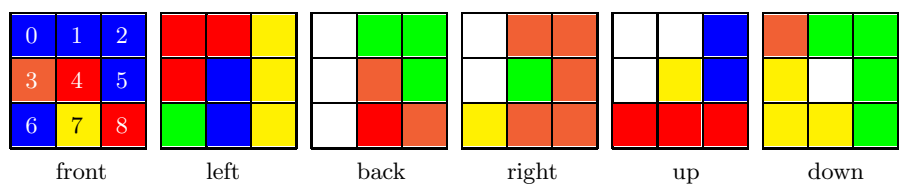
\begin{figure}
\setlength{\unitlength}{1mm}
%\psset{unit=1mm}
\begin{picture}(120,33)

 %\put(0,36.6){\framebox(7,7){\color{red} 1}}
\multiput(0,10)(6,0){4}{\line(0,1){18}}
\multiput(0,10)(0,6){4}{\line(1,0){18}}

\multiput(0.3,22.4)(0.1,0){55}{\color{blue} \line(0,1){5.3}}
\multiput(6.3,22.4)(0.1,0){55}{\color{blue} \line(0,1){5.3}}
\multiput(12.3,22.4)(0.1,0){55}{\color{blue} \line(0,1){5.3}}

\multiput(0.3,16.4)(0.1,0){55}{\color{RedOrange} \line(0,1){5.35}}
\multiput(6.3,16.4)(0.1,0){55}{\color{red} \line(0,1){5.35}}
\multiput(12.3,16.4)(0.1,0){55}{\color{blue} \line(0,1){5.35}}

\multiput(0.3,10.4)(0.1,0){55}{\color{blue} \line(0,1){5.4}}
\multiput(6.3,10.4)(0.1,0){55}{\color{yellow} \line(0,1){5.4}}
\multiput(12.3,10.4)(0.1,0){55}{\color{red} \line(0,1){5.4}}

\put(2,24){\color{white} 0} \put(8,24){\color{white}1}
\put(14,24){\color{white} 2}

\put(2,18){\color{white} 3} \put(8,18){\color{white} 4}
\put(14,18){\color{white} 5}

\put(2,12){\color{white} 6} \put(8,12){\color{black} 7}
\put(14,12){\color{white} 8}

%{\color{blue}\linethickness{5.6mm}\put(0.22,13.02){\line(1,0){5.6}}}
%{\color{blue}\linethickness{5.6mm}\put(5.3,13.02){\line(1,0){5.3}}}
%{\color{blue}\linethickness{5.6mm}\put(10.2,13.02){\line(1,0){5.35}}}

\multiput(20,10)(6,0){4}{\line(0,1){18}}
\multiput(20,10)(0,6){4}{\line(1,0){18}}

\multiput(20.3,22.4)(0.1,0){55}{\color{red} \line(0,1){5.3}}
\multiput(26.3,22.4)(0.1,0){55}{\color{red} \line(0,1){5.3}}
\multiput(32.3,22.4)(0.1,0){55}{\color{yellow} \line(0,1){5.3}}

\multiput(20.3,16.4)(0.1,0){55}{\color{red} \line(0,1){5.35}}
\multiput(26.3,16.4)(0.1,0){55}{\color{blue} \line(0,1){5.35}}
\multiput(32.3,16.4)(0.1,0){55}{\color{yellow} \line(0,1){5.35}}

\multiput(20.3,10.4)(0.1,0){55}{\color{green} \line(0,1){5.4}}
\multiput(26.3,10.4)(0.1,0){55}{\color{blue} \line(0,1){5.4}}
\multiput(32.3,10.4)(0.1,0){55}{\color{yellow} \line(0,1){5.4}}

\multiput(40,10)(6,0){4}{\line(0,1){18}}
\multiput(40,10)(0,6){4}{\line(1,0){18}}

\multiput(46.3,22.4)(0.1,0){55}{\color{green} \line(0,1){5.3}}
\multiput(52.3,22.4)(0.1,0){55}{\color{green} \line(0,1){5.3}}

\multiput(46.3,16.4)(0.1,0){55}{\color{RedOrange} \line(0,1){5.35}}
\multiput(52.3,16.4)(0.1,0){55}{\color{green} \line(0,1){5.35}}

\multiput(46.3,10.4)(0.1,0){55}{\color{red} \line(0,1){5.4}}
\multiput(52.3,10.4)(0.1,0){55}{\color{RedOrange} \line(0,1){5.4}}

\multiput(60,10)(6,0){4}{\line(0,1){18}}
\multiput(60,10)(0,6){4}{\line(1,0){18}}

\multiput(66.3,22.4)(0.1,0){55}{\color{RedOrange} \line(0,1){5.3}}
\multiput(72.3,22.4)(0.1,0){55}{\color{RedOrange} \line(0,1){5.3}}

\multiput(66.3,16.4)(0.1,0){55}{\color{green} \line(0,1){5.35}}
\multiput(72.3,16.4)(0.1,0){55}{\color{RedOrange} \line(0,1){5.35}}

\multiput(60.3,10.4)(0.1,0){55}{\color{yellow} \line(0,1){5.4}}
\multiput(66.3,10.4)(0.1,0){55}{\color{RedOrange} \line(0,1){5.4}}
\multiput(72.3,10.4)(0.1,0){55}{\color{RedOrange} \line(0,1){5.4}}

\multiput(80,10)(6,0){4}{\line(0,1){18}}
\multiput(80,10)(0,6){4}{\line(1,0){18}}

\multiput(92.3,22.4)(0.1,0){55}{\color{blue} \line(0,1){5.3}}

\multiput(86.3,16.4)(0.1,0){55}{\color{yellow} \line(0,1){5.35}}
\multiput(92.3,16.4)(0.1,0){55}{\color{blue} \line(0,1){5.35}}

\multiput(80.3,10.4)(0.1,0){55}{\color{red} \line(0,1){5.4}}
\multiput(86.3,10.4)(0.1,0){55}{\color{red} \line(0,1){5.4}}
\multiput(92.3,10.4)(0.1,0){55}{\color{red} \line(0,1){5.4}}

\multiput(100,10)(6,0){4}{\line(0,1){18}}
\multiput(100,10)(0,6){4}{\line(1,0){18}}

\multiput(100.3,22.4)(0.1,0){55}{\color{RedOrange} \line(0,1){5.3}}
\multiput(106.3,22.4)(0.1,0){55}{\color{green} \line(0,1){5.3}}
\multiput(112.3,22.4)(0.1,0){55}{\color{green} \line(0,1){5.3}}

\multiput(100.3,16.4)(0.1,0){55}{\color{yellow} \line(0,1){5.35}}
\multiput(112.3,16.4)(0.1,0){55}{\color{green} \line(0,1){5.35}}

\multiput(100.3,10.4)(0.1,0){55}{\color{yellow} \line(0,1){5.4}}
\multiput(106.3,10.4)(0.1,0){55}{\color{yellow} \line(0,1){5.4}}
\multiput(112.3,10.4)(0.1,0){55}{\color{green} \line(0,1){5.4}}

\put(6,6){front} \put(26,6){left} \put(46,6){back} \put(66,6){right}
\put(88,6){up} \put(106,6){down}

\end{picture}
\caption{ A state of Rubik's Cube} \label{sketchFig}
\end{figure}

Rubik's Cube is a 3-D mechanical cube, which consists of 27 smaller
cubes (cubies). The center cubies on each face and the core of
Rubik's Cube forms a fixed frame. Other 20 cubies move around them.
A full face of the larger cube is divided into 9 facelets, each of
which is a face of a distinct cubies, where each face of the cubies
is colored one of six colors. A state of Rubik's Cube can be
considered as a permutation on 48 facelets, since 6 center facelets
are fixed. In general, it may be described by six faces: ``front",
``left", ``back", ``right", ``up" and ``down" face, each with
$3\times 3$ facelets.  Figure 1 presents a sate (position) of
Rubik's Cube. A state of Rubik's Cube is said to be the home state
(position) or solved state (position) if all facelets of each face
in that state are the same color. Solving Rubik's Cube means
restoring from a scrambled state into the home state.

A move on Rubik's Cube refers to rotating the nine cubies on a face
as a group 90 or 180 degrees around a central axis. We use the `face
turn metric' to compute the number of moves required to solve
Rubik's Cube. That is, a single move is considered as a turn of any
face, 90 or 180 degrees in any direction.

Rubik's Cube has a total of eighteen different moves. These moves
are conventionally denoted by $U, U^\prime, U2, D, D^\prime, D2, L,
L^\prime, L2, R, R^\prime, R2, F, F^\prime, F2, B, B^\prime$, and
$B2$. Each clockwise 90 degree move is specified by just the face
with no suffix, and each counterclockwise 90 degree move and each
180 degree move are specified by the face followed by a prime symbol
($^\prime$), and the digit 2, respectively. So $U$ here denotes a
clockwise quarter turn of the ``up" face, and similarly, $D, L, R,
F$ and $B$ denote ``down", ``left", ``right", ``front" and ``back",
respectively. A solution can be represented by a move sequence. As
an example, the move sequence
$F2U2B^{\prime}U^{\prime}B2D^{\prime}U2F{^\prime}U2LDR2B2U2F^{\prime}U2F{^\prime}U2B2L2$
is a solution to the state shown in Figure 1. That is, performing in
turn each move in this sequence can restores that state to the home
state. As defined in \cite{Rokicki:3}, we define $S_{18}$ the set of
18 moves mentioned above, and $A_{10} = \{U, U^\prime, U2, D,
D^\prime, D2, L2, R2, F2, B2\}$, which will be used in subsequent
sections and the two-phase algorithm given later.

A Rubik's Cube consists of different cubies. By convention, the
cubies are classified into \emph{edge cubies} (two visible
facelets), \emph{corner cubies} (three visible facelets) and
\emph{center cubies} (one visible facelets, in the center of a
side). Correspondingly, according to the cubie where a facelet
belongs, the facelets are classified into \emph{edge facelets},
\emph{corner facelets} and \emph{center facelets}.

One of goals of this paper is to translate the Rubik's cube puzzle
into a satisfability (SAT) problem with CNF (conjunctive normal
form). A propositional logic formula is said to be in CNF if it is a
conjunction (``and") of clauses, each clause being a disjunction
(``ors") of Boolean literals, where each literal is either a
variable or the negation of a variable.

\section{SAT encoding of the Rubik's Cube puzzle}

The Rubik's Cube puzzle may be described by the initial state, the
move sequence, the map relation of each move and the solved state.
Its SAT encoding will consist of such ingredients. A Rubik's Cube
has a total of six colors. A naive approach is that a color
corresponds a Boolean variable. Thus, representing each color on
each facelet requires six Boolean variables. In fact, six colors
contains only $\log 6\approx 2.6 $ bit information. So the number of
Boolean variables can be reduced. Let $b_1b_2b_3$ be the binary
representation of $k (0\leq k \leq 5)$. The Boolean variable
representation of the $k$-th color is $x_1(b_1),x_2(b_2),x_3(b_3)$,
where $x_i(b_i)$ is $x_i$ if $b_i$ is 1, and $\overline{x_i}$
otherwise. For example, the Boolean variable representation of the
second color is $\overline{x_1},x_2,\overline{x_3}$. Therefore, 3
Boolean variables suffice for representing the color of each
facelet. In our SAT encoding, states are divided into two
categories: general state and $H$-state. A state is said to be
$H$-state if it can be transformed into the solved state by a
sequence of the moves in $A_{10}$ mentioned above. In the two-phase
algorithm, each state in Phase two is $H$-state. For general states,
we represent each color on each facelet with three Boolean
variables. For H-states, we represent each color on each facelet
with two Boolean variables. In the 2-variable scheme, we represent
the colors of the front, left, back, right face in the solved state
by 00, 01, 10 and 11, respectively, and then re-use 00 and 01 to
represent the colors of  the other two (top and down) faces. Notice,
any move in in $A_{10}$ cannot transform any facelet on top and down
faces to somewhere on the other four faces. $H$-states are allowed
to use only moves in $A_{10}$. Therefore, under $H$-states, the
2-variable scheme does not yield any confusing.

Let $c(i, j, m)$ be the color of the $j$-th facelet in the $i$-the
face under the $m$-th ($m\geq 1$) state (hereafter, color of facelet
$(i,j,m)$, for short), $c(i,4,1)$ the center facelet color of the
$i$-th face under the initial state. If the $m$-th state is the
solved state, this state may be represented by

\hskip 10mm $\bigwedge\limits_{1\leq i \leq 6, 0 \leq j \leq 8}
c(i,j,m)=c(i,4,1)$\\
Using 3-variable scheme, $c(i, j, m) = c(i,4,1)$ is translated into
$c(i,j,m,1) = c(i,4,1,1) \wedge c(i,j, m,2) = c(i,4,1,2) \wedge
c(i,j, m,3) = c(i,4,1,3)$, where $c(\ldots 1)$, $c(\ldots 2)$ and
$c(\ldots 3)$ are literals that denote the 1st, 2nd and 3rd bit of a
color. Formula $c(i, j, m,1) = c(i,4,1,1)$ can be translated into
the following clauses: $(c(i, j, m, 1)\vee \neg c(i, 4, 1,1)) \wedge
( \neg c(i, j, m,1) \vee c(i,4,1,1))$.

An initial state of a cube is considered as State 1, which is
interpreted as

\hskip 10mm $\bigwedge\limits_{1\leq i \leq 6, 0 \leq j \leq 8,
k=1,2,3} B(c(i,j,1,k))$\\
where $B(c(i,j,1,k))$ is defined as $c(i,j,1,k)$ if the value of the
$k$-th bit color of facelet$(i,j,1)$ is 1, and $\neg c(i,j,1,k)$
otherwise.

Assume we take at most $n-1$ moves to solve Rubik's Cube, and
associate a Boolean variable st with each state $t (1\leq t \leq
n)$.  `` $s_t =$ true " mean the $t$-th state is the solved state.
Then, this constraint can be represented by

\hskip 10mm $\bigwedge\limits_{1\leq i \leq 6, 0 \leq j \leq 8}
( \neg s_t \vee c(i,j,t)=c(i,4,1))$\\
This formula can be converted easily into clauses.

At any time, among $S=\{s_1, s_2,\ldots,s_n \}$, we must ensure that
exactly one $s_t$ is true. The \emph{exactly-one} constraint can be
formalized by the \emph{at-least-one} (ALO) and \emph{at-most-one}
(AMO) constraint. That is, exactly-one$(S) \equiv $ ALO $(S)\wedge $
AMO $(S)$. The standard SAT encodings of constraints ALO and AMO are
the following.

\hskip 10mm $\mathrm{ALO}(S)\equiv s_1 \vee s_2 \vee \cdots \vee
s_n$

\hskip 10mm $\mathrm{AMO}(S)\equiv \{ \overline{s_i}\vee
\overline{s_j}|s_i,s_j\in S, i<j\}$

\noindent The ALO constraint ensures that a variable is true. And
the AMO constraint ensures that no more than one variable is true.
The standard AMO encoding requires much more clauses. To reduce the
number of clauses, we can apply a two-product AMO encoding
\cite{chenAMO:14}, which is recursively defined as

$\mathrm{AMO}(S)\equiv \mathrm{AMO}(U)\wedge \mathrm{AMO}(V)
\bigwedge\limits_{1\leq i \leq p,1\leq j \leq q}^{ 1 \leq k \leq n,
k=(i-1)q+j}((\overline{x_k}\vee
u_i)\wedge (\overline{x_k}\vee v_j))$\\
where $ p=\lceil \sqrt{n} \rceil, q=\lceil \frac{n}{p} \rceil,
U=\{u_1,u_2,\ldots,u_p\},V=\{v_1,v_2,\ldots,v_q\}$, each element
$u_i$ in $U$ and each element $v_j$ in $V$ are auxiliary variables.
Here, $\mathrm {AMO} (U)$ and $\mathrm {AMO} (V)$ apply the standard
AMO encoding. The number of clauses and auxiliary variables required
by this encoding are $2n+p(p-1)/2+q(q-1)/2$ and $p+q$, respectively.
When $n=20$, the number of clauses required is
$40+4(4-1)/2+5(5-1)/2=56$. For the $n$, the standard AMO encoding
requires 20(20-1)/2=190 clauses.

To encode efficiently the constraints on the turns, we classify the
turns of Rubik's Cube into six classes: $u, d, l, r, f$ and $b$. Let
$u_k (1\leq k\leq n)$ be a Boolean variable that is associated with
the up turn of step $k$. We perform either $U$-, or $U^\prime$- or
$U2$-type up turn at step $k$ when $u_k$ is true, and do the other
turn otherwise. The meaning of $d_k, l_k, r_k, f_k$ and $b_k$ is
similar. At any step, we have a unique turn. This constraint can be
formalized by exactly-one$(u_k, d_k, l_k, r_k, f_k ,b_k)$ for $1\leq
k \leq n$. Each $u_k$ corresponds actually three different turn: $U,
U^\prime, U2$. We denote the $U, U^\prime, U2$ of step $k$ by
Boolean variables $U_k, U_k^\prime, U_{k}2$. Clearly, these Boolean
variables should satisfy $\neg u_k \vee $ exactly-one($U_k,
U_k^\prime, U_{k}2$). Similarly, we have the following constraint
conditions:

$\neg d_k \vee $ exactly-one($D_k, D_k^\prime, D_{k}2$)

$\neg l_k \vee $ exactly-one($L_k, L_k^\prime, L_{k}2$)

$\neg r_k \vee $ exactly-one($R_k, R_k^\prime, R_{k}2$)

$\neg f_k \vee $ exactly-one($F_k, F_k^\prime, F_{k}2$) and

$\neg b_k \vee $ exactly-one($B_k, B_k^\prime, B_{k}2$).

A move can be considered as a mapping that maps each facelet $c(i,
j, k) (1\leq i \leq 6, 0\leq j \leq 8, 1 \leq k \leq n)$ at State
$k$ to a facelet $c(i^\prime, j^\prime, k-1)$ at State $k-1$. Let
$M_k$ be a Boolean variable denoting one of 18 different moves at
step $k$. The corresponding mapping is denoted by $f_{M_k}$. Then we
have the following constraint condition.

\hskip 10mm $ \neg M_k \vee \bigwedge\limits_{1\leq i \leq 6, 0 \leq
j \leq 8} c(i,j,k)=f_{M_k}(c(i,j,k))$ \\
For the clockwise up turn $U_k$, the mapping relationship of
$f_{U_k}$: $c(i,j,k)\rightarrow c(i^\prime,j^\prime,k-1)$ is the
following.

$c(i$ $\mathrm{mod}$ $4+1,0,k)= c(i,0,k-1),c(i$ $\mathrm{mod}$
$4+1,1,k)= c(i,1,k-1)$,

$c(i$ $\mathrm{mod}$ $4+1,2,k)= c(i,2,k-1)$ for $1\leq i\leq 4 $

$c(5,0,k)= c(5,6,k-1)$, $c(5,1,k)= c(5,3,k-1)$,

$c(5,2,k)= c(5,0,k-1)$, $c(5,3,k)= c(5,7,k-1)$,

$c(5,5,k)= c(5,1,k-1)$, $c(5,6,k)= c(5,8,k-1)$,

$c(5,7,k)= c(5,5,k-1)$, $c(5,8,k)= c(5,2,k-1)$\\
The other facelets keep unchanged.

It is easy to see that any of the moves will move exactly 20
facelets, and preserve the other facelets. Furthermore, in the same
type of moves, the unchanged facelets are the same. For example, for
$U$, $U^\prime$, $U2$, they all preserve the facelets 3-8 in faces
1-4, and all the facelets of face 6. To save the number of clauses,
we split the above constraint condition into two parts: changed and
unchanged. We use Boolean variables denoting the same type of moves
to control the unchanged part. The changed part is controlled by
Boolean variables denoting concrete moves. For the move $U$ of step
$k$, we have the following constraint conditions.

$ \neg u_k \vee \bigwedge\limits_{0 \leq j \leq 8}
c(6,j,k)=c(6,j,k-1) \bigwedge\limits_{1\leq i \leq 4, 3 \leq j \leq
8} c(i,j,k)=c(i,j,k-1)$

$ \neg U_k \vee \bigwedge\limits_{0 \leq j \leq 8}
c(6,j,k)=c(6,f(j),k-1) \bigwedge\limits_{1\leq i \leq 4, 0 \leq j
\leq 2} c(g(i),j,k)=c(i,j,k-1)$\\
where $f(j)$ is the mapping as shown above, and $g(i) = i$ mod $4
+1$. For the move $U^\prime$ ($U2$) of step $k$, the first condition
above will share. The second condition needs to re-construct, but is
easy. This can be done by replacing $U_k$ in the second condition
above with $U_k^\prime$ ($U_k2$), and defining the corresponding
$f(j)$ and $g(i)$. For the other moves such as $D, F, B, L, R$,
similar constraint conditions are easily constructed. The
optimization technique given above can reduce the number of clauses
by above 1/3.

Some two-move sequences will yield the same result. For example,
two-moves $UD$ and $DU$ have the same result states. To speed up the
search, we remove the search on two-move sequences such as $DU$. The
removing of such a search can be done by adding the following
constraint clauses to the SAT encoding of Rubik's Cube.

$ \bigwedge\limits_{1 \leq k \leq n} ((\overline{u_k} \vee
\overline{d_{k+1}}) \wedge (\overline{l_k} \vee \overline{r_{k+1}})
\wedge (\overline{f_k} \vee \overline{b_{k+1}})) $

\section{Encoding Kociemba's algorithm and other constraints}

Using the SAT encoding given in the previous section, a modern SAT
solver can find only solutions of length at most 13. However, many
states have already been shown that requires 20 moves (e.g.
superflip). To find such a solution, we add other tricks. A useful
trick is to add the encoding of  Kociemba's algorithm, which is a
two-phase algorithm. The basic idea of the algorithm is to splits
the problem into two almost equal subproblems, each of which can use
a lookup table to search for exhaustively a solution. Here is the
pseudo-code of Kociemba's algorithm.

\begin{flushleft}
{\bf Kociemba's Algorithm} \\
\hskip 4mm $d \leftarrow 0$,  $t \leftarrow \infty$\\
\hskip 4mm {\bf while} $d < t$ {\bf do} \\
\hskip 8mm    {\bf for} $s \in S_{18}^d$, $ps \in H$ {\bf do}\\
\hskip 12mm        {\bf if} $d+D(ps) < t$ {\bf then} \\
\hskip 16mm             find a better solution, using moves in $A_{10}$ \\
\hskip 16mm             $ t \leftarrow d+D(ps)$ \\
\hskip 12mm         {\bf end if} \\
\hskip 8mm     {\bf end for} \\
\hskip 8mm     $ d \leftarrow d+1$ \\
\hskip 4mm     {\bf end while} \\
\end{flushleft}

This algorithm assumes that the original state is $p$, and applies
some move sequence $s \in S_{18}^d$  (for the definition of
$S_{18}$, see Section 2) of length $d$ to the original cube yielding
$ps$ which lies in $H$. This search process is called phase one.
Here $H$ is a subset of states that is composed of all patterns with
following characteristics:

\begin{enumerate}
\item The orientation of all corner cubies and edge cubies is correct.

\item The edge cubies that should be in the middle layer are now located
in the middle layer.

\end{enumerate}

These characteristics are preserved by moves in the set $A_{10}$
(which is defined in Section 2). The search process from the new
state $ps$ to the fully solved state is called Phase two. In this
phase, each move is in $A_{10}$. $D(ps)$ returns the distance from
the state $ps$ to the solved state using moves in $A_{10}$. To
efficiently complete this computation, it is usually done by a
lookup table. In fact, Phase one is usually also done by a lookup
table.

It is impossible to encode directly the entire Kociemba's algorithm
into a SAT problem, because it contains lookup tables. However, it
is possible to encode the basic idea of Kociemba's algorithm with a
CNF formula. Let Cube\_CNF($n$) denote a SAT encoding of Rubik's
Cube with a total of $n$ states, which can easily be done by the
approach given in Section 3. Assume that the $k$-th state $s_k$
reaches a state in $H$. A SAT encoding of Rubik's Cube containing
the basic idea of Kociemba's algorithm can be described by the
following formula:

Cube\_CNF($n$) $\wedge$ Hstate($s_k$) $\wedge$ $A_{10}$\_move($k,n$) \\
where Hstate($s_k$) is true if $s_k$ is in $H$, and
$A_{10}$\_move($k,n$) is used to restrict  moves from step $k$ to
step $n$ to be  moves in $A_{10}$. Hstate($s_k$) is defined as

$ \bigwedge\limits_{1\leq i \leq 4, j=3,5} (c(i,j,k)=c(i,j,4) \vee
c(p(i),j,k)=c(i,j,4)) \wedge $

$ \bigwedge\limits_{i=5,6 \wedge 0\leq j \leq 8} (c(i,j,k)=c(i,j,4)
\vee c(p(i),j,k)=c(i,j,4)) $ \\
where $p(i)$ denotes the opposite face of the $i$-th face, i.e., the
mapping relationship of $p$ is: $1 \leftrightarrow 3$,  $2
\leftrightarrow 4$, $5 \leftrightarrow 6$.

Based on the definition of $A_{10}$, $L, L^\prime, R, R^\prime,F,
F^\prime, B$ and $B^\prime $ all are not in  $A_{10}$. So
$A_{10}$\_move($k, n$) may be described by the following logic
formula.

$ \bigwedge\limits_{k < m \leq n} \neg (L_m \vee L_m^\prime \vee R_m
\vee R_m^\prime \vee F_m \vee F_m^\prime \vee B_m \vee B_m^\prime)$

This means that after step $k$, neither clockwise nor counter
clockwise 90 degree turn of any face except for the up and down face
is allowed.

The encoding of Rubik's Cube containing the above two constraints
can be considered as a SAT encoding of Kociemba's algorithm.
Depending on different $k$, the efficiency of solution is different.
In general, $k$ is set to less than 12. That is, the length of Phase
one is limited to 12.

A move on a cube can change only 20 facelets, and keep the other 28
facelets unchanged. That is to say, the last second state from the
solved state have 28 facelets that are placed correctly. Based on
this property, we encode the last second state as a additional
constraint condition. Let $t$ is the last turn operation. Then we
add the following encodings:

$(\neg u_t \vee $ unchanged28facelet($U$)) $\wedge (\neg d_t \vee $
unchanged28facelet($D$)) $\wedge$

$(\neg l_t \vee $ unchanged28facelet($L$)) $\wedge (\neg r_t \vee $
unchanged28facelet($R$)) $\wedge$

$(\neg f_t \vee $ unchanged28facelet($F$)) $\wedge (\neg b_t \vee $
unchanged28facelet($B$)).

\noindent where unchanged28facelet($U$) can be encoded as follows.

\hskip 10mm $\bigwedge\limits_{1\leq i \leq 4, 3 \leq j \leq 8}
c(i,j,t-1)=c(i,4,1) \bigwedge\limits_{0 \leq j \leq 8}
c(6,j,t-1)=c(6,4,1)$ \\
the other unchanged28facelets are similar. The last second state
from the final state of phase 1 can be encoded also in a similar
way.

\section{A SAT Solver for Solving Rubik's Cube }

Based on our experimental observation, the PrecoSAT solver
\cite{Precosat:13}, the Gold Medal winners in the application
category of the SAT 2009 competition, was the fastest on Rubik's
Cube. Without any pruning strategy, it is hard to solve Rubik's
Cube. Since PrecoSAT has no pruning strategy, it is not good choice
to use directly PrecoSAT. To solve more efficiently Rubik's Cube, we
built a new solver based on PrecoSAT. The basic framework of this
new solver is similar to MoRsat \cite {MoRsat:15}, but replaces the
lookahead solving strategy with an ALO (\emph{at-least-one}) solving
strategy. Let the notation ${\cal F}(x)$ denotes the resulting
formula after assigning literal $x$ true and performing iterative
unit propagation. The basic idea of the ALO solving strategy is to
decompose the original problem ${\cal F}$ into subproblems ${\cal
F}(x_i) (1\leq i \leq n) $ if a clause $C$ in ${\cal F}$ is $x_1
\vee x_2 \vee \cdots \vee x_n$. If no subproblem is satisfiable, the
original problem is unsatisfiable. Each subproblem can be solved in
a recursive way. Once the recursive depth reaches some constant, say
4, we use PrecoSAT to solve that subproblem. This ALO solving
strategy has been applied successfully to MPhaseSAT
\cite{MPhaseSAT:17}. In the new solver used for Rubik's Cube, the
size of the clause $C$ used to decompose the original problem is
specified to 6. And the recursive depth of the ALO solving strategy
is limited to 4. Under this assumption, this new solver may be
described in a recursive way as follows.

\begin{flushleft}
{\bf Algorithm } SATsolver(${\cal F}, level$)\{Initially $level$ is set to 1\} \\
\hskip 4mm    ${\cal F} \leftarrow $ LookaheadSimplify($ \cal F$)\\
\hskip 4mm find a clause $C$ with 6 free variables\\
\hskip 4mm {\bf if } no such $C$ was found {\bf then} {\bf return} PrecoSAT(${\cal F}$)\\
\hskip 4mm {\bf for} $i=1$ {\bf to} 6 {\bf do}\\
\hskip 12mm    assume $C =x_1 \vee x_2 \vee x_3 \vee x_4 \vee x_5 \vee x_6$ \\
\hskip 12mm    ${\cal F}' \leftarrow {\cal F}(x_i)$ \\
\hskip 12mm    {\bf if} $level \leq 4$ {\bf then}\\
\hskip 20mm              {\bf if} SATsolver($ {\cal F}', level+1) =  \mathrm {SAT}$ {\bf then return} satisfiable \\
\hskip 12mm    {\bf else} {\bf if} PrecoSAT($ {\cal F}') = \mathrm {SAT}$ {\bf then return} satisfiable \\
\hskip 4mm {\bf end for}\\
\hskip 4mm {\bf return } unsatisfiable
\end{flushleft}

Procedure LookaheadSimplify corresponds multiple failed literal
probes in PrecoSAT. It can be either a simple look-ahead or double
look-ahead procedure in March \cite {March:18}. In some cases, this
procedure can be removed. The reason why the size of the clause $C$
in the above algorithm is limited to 6 is that there are six types
of moves, and in the SAT encoding of Rubik's Cube, the clause of
size 6 is certainly obtained by encoding six types of moves. For
general problems, the size of the clause $C$ in the above algorithm
should be as long as possible.

\section{Experimental Studies}

We solved Rubik's Cube with the SAT solver described in previous
section on a 2.40GHz machine with Intel Core 2 Quad Q6600 CPU.
Without adding the constraint conditions of two-phase algorithm
given in Section 4, we found that determining whether a cube has an
optimal solution with the maneuver (a move sequence is called a
maneuver) length of 13 took about 7 hours in the worst case by our
SAT solver. If the maneuver length of a solution is greater than 13,
in general, no modern SAT Solver cannot find efficiently an optimal
solution. However, if adding the constraint condition of two-phase
algorithm, it is easy to find a near-optimal solution. Hence, as a
part of the SAT encoding of a cube, all the subsequent experiments
assume that constraint condition of two-phase algorithm is add to
the SAT encoding of Rubik's Cube. Table\,1 presents the numbers of
variables and clauses required to encode Rubik's Cube by our
encoding strategy. Here, the length of a solution is limited to be
20. So all the numbers of clauses required  are small, and are
within 6700, although they have a little bit change for different
lengths of the phase 1 maneuvers.

\begin{table}
\caption{ Numbers of variables and clauses required to encode
Rubik's Cube with different lengths of the phase 1 maneuvers}
\begin{center}

\setlength\tabcolsep{4pt}
%\begin{tabular}{p{4.3cm}|c|c|c|c}
\begin{tabular}{c|c|c}
\hline  \hline
%\rule{\textwidth}{1pt}\\
\multicolumn{1}{c|} {length of phase 1} & number of variable  & number of clauses \\
\hline

9  & 3570 & 66028  \\
10 & 3570 & 66026  \\
11 & 3594 & 66201  \\
12 & 3618 & 66248  \\

\hline
\end{tabular}
\end{center}
\end{table}

The time to solve a cube greatly depends on the given states. The
cube state shown in Figure 1 is an easy example. We found the
following six solutions to this example by setting different faces
to the $U$- and $D$-face, and using different encoding strategies,
based on two-phase algorithm.

\begin{enumerate}
\item  $F2RDF^{\prime}R^{\prime}D2UL^{\prime}DFL2UR2D2L2U^{\prime}L2B2F2R2$

\item $R^{\prime}FRBU^{\prime}B^{\prime}D^{\prime}LU^{\prime}F^{\prime}U^{\prime}R2B2U^{\prime}B2L2U2L2D2F2$

\item $F^{\prime}L^{\prime}F^{\prime}U2BU^{\prime}F^{\prime}DF2LB2R2D2F2R2FD2B^{\prime}F2U2$

\item $UF^{\prime}RL^{\prime}B2F2LBD^{\prime}B^{\prime}RB2LB2LD2B2L^{\prime}B2U2$

\item
$UF^{\prime}D2U2R^{\prime}L2FU^{\prime}F^{\prime}RF2R^{\prime}L2D2LF2D2R^{\prime}B2U2$

\item $UFR2LB2F2L^{\prime}B^{\prime}U^{\prime}B^{\prime}RD2B2D2B2U2L2B2R^{\prime}L^{\prime}$

\end{enumerate}

Each solution took about 200 seconds. These solutions have a common
characteristic: both the length of the phase 1 maneuvers and the
length of the phase 2 maneuvers are 10, and the total length is 20.
These solutions are not the shortest. Finding a shorter solution
will take much more time, since the length of the phase 1 maneuvers
increases with the shortening of solutions.

A special cube state which flips all 12 edges, called superflip, is
a hard example to our SAT solver. It has been proven to have a
shortest maneuver length of 20 moves to be solved. Finding a
shortest solution to the superflip is time-consuming. However, it is
easy to find a near-optimal solution. Actually, we took 532 seconds
to find a solution with the length of 21 moves as follows.

\hskip 10mm $BF^{\prime}L^{\prime}U2F2L
D^{\prime}U^{\prime}F^{\prime}R^{\prime}LF2U2R2B2UR2D^{\prime}B2U^{\prime}R2$

If the length of the phase 1 maneuvers is known to be 13 in advance,
for the superflip, we can find easily a shortest solution of length
20 as follows.

\hskip 10mm
$BFU2R^{\prime}D^{\prime}UL2B2R2B^{\prime}U2R^{\prime}L^{\prime}U^{\prime}L2U^{\prime}B2D^{\prime}L2U^{\prime}$

It took about 310 seconds. Note that this solution is different from
that given by Cube Explorer \cite{Explorer:16} that implements
two-phase algorithm using lookup tables.

To test the generalized case, we generated 10 randomly cube states.
For each state, the time required by our SAT solver to search for a
solution of length 20 or less is shown in Table\,2.

The upper bound on the length of the phase 1 maneuvers has been
shown to 12 \cite{Reid:7}. Our experiments verified that this fact
is true. Furthermore, within length 12 of the phase 1 maneuvers, we
found always a solution of length 20 or less. As the length of the
phase 1 maneuvers increases, the time to solve a cube increases
sharply. In most of the cases, we can find a solution within 7000
seconds. If the length of solutions is allowed to be 21, the time to
solve a cube by our SAT solver never exceeds 1500 seconds.

\begin{table}
\caption{Runtime took by our SAT solver to solve 10 random cube
states}
\begin{center}

\setlength\tabcolsep{4pt}
%\begin{tabular}{p{4.3cm}|c|c|c|c}
\begin{tabular}{c|c|c|c}
\hline  \hline
%\rule{\textwidth}{1pt}\\
\multicolumn{1}{c|}{cube state} & length of phase 1 & length of phase 2 & time (seconds)\\
\hline

1 &  10 & 10 & 1072 \\
2 &  11 & 9  & 6785 \\
3 &  10 & 10 & 137 \\
4 &  11 & 9  & 6123 \\
5 &  9  & 11 &  87 \\
6 &  10 & 10 & 774 \\
7 &  11 & 9  & 1489 \\
8 &  10 & 9  & 1079 \\
9 &  12 & 8  & 14096 \\
10 & 10 & 10 & 329 \\
\hline
\end{tabular}
\end{center}
\end{table}

\begin{table}
\caption{Runtime took by Cube Explorer to solve 10 random cube
states}
\begin{center}

\setlength\tabcolsep{4pt}
%\begin{tabular}{p{4.3cm}|c|c|c|c|c}
\begin{tabular}{c|c|c|c|c}
\hline  \hline
%\rule{\textwidth}{1pt}\\
\multicolumn{1}{c|}{cube state} & solution length & time (seconds) & solution length & time (seconds)\\
 & (non-optimal) & (non-optimal) & (optimal) & (optimal) \\
\hline

1  & 19 & 0.3 & 19 & 9125 \\
2  & 19 & 4.5 & 18 & 531 \\
3  & 18 & 1.5 & 18 & 580 \\
4  & 16 & 7.3 & 16 & 23 \\
5  & 19 & 0.3 & 18 &  3243 \\
6  & 18 & 1.5 & 17 &  315 \\
7  & 17 & 2.1 & 17 & 25 \\
8  & 19 & 0.1 & 18 & 4216 \\
9  & 19 & 1.1 & 18 & 2219 \\
10 & 19 & 0.1 & 18 &  1452 \\

\hline
\end{tabular}
\end{center}
\end{table}

 The test platform in Table\,3 are  the same as that in
 Table\,2. Compared with the SAT solver, Cube Explorer using lookup tables is much faster.
 As shown in Table\,3, Cube Explorer took at most 8 seconds to find a non-optimal
 solution whose length is at most 20. In most of the cases, it found
 a solution immediately. However, to find an optimal
 solution, it is also time-consuming in some cases. For example, finding an
 optimal solution of cube state 1 took 9125 seconds. If information on lookup
 tables can be encoded in CNF, it is possible that the SAT solver
 can outperform Cube Explorer.

  Our encoding follows almost the idea of brute force
enumeration. However, it is impossible for brute force enumeration
to find a solution of length 20 with length 12 of the phase 1
maneuvers on a modern PC in a reasonable time, since brute force
enumeration has to check about $15^{12}\times 9^{10} \approx
10^{23.7}$ states (note each state consists of 48 movable facclets)
in the worst case. Hence, the SAT solver has its advantage over
generalized approaches such as brute force enumeration.

  The main advantage of solving a cube with SAT solvers is that it
does not need a huge lookup table indicating the distance from the
home state, and the memory requirement is very less. 10 MB RAM
memory is enough.

\section {Conclusions}

This paper is the first to solve Rubik's Cube using a SAT solver.
The experimental results reveal that our SAT encoding of Rubik's
Cube and the improvement on the existing SAT solver are effective.
Using the improved SAT solver, Rubik's Cube can be solved in a
reasonable time. We believe that the encoding approaches and the ALO
solving technique should be useful beyond the planning domain such
as Rubik's Cube.

Here many open problem remains. For example, what is the optimal SAT
encoding of Rubik's Cube?  The heuristic approach is frequently used
to speed up SAT solvers. Can the heuristics information be encoded
into a SAT formula?  Indeed, in this paper, we implemented the SAT
encoding of partial heuristic information such as the goal state
information of phase 1 in two-phase algorithm. However, we cannot
still encode the information on lookup tables (pattern database of
Rubik's Cube) in Kociemba's algorithm, which is used to prune the
superfluous search space. When a SAT formula is given, can we
exploit logic structures on the heuristic information? What
two-phase algorithm uses is a depth-first search technique. How to
encode the depth-first search technique in CNF is also a challenge.
Although one now has proved that God's number or the upper bound on
the number of moves for Rubik's Cube is exactly 20, We cannot encode
yet such a problem into a SAT problem. Nevertheless, in near future,
we believe that it is possible to perform the SAT encoding of the
upper bound of Rubik's Cube by extending the current SAT encoding
technique of Rubik's Cube.

\bibliographystyle{splncs}
\bibliography{Rubikcube}

\begin{thebibliography}{[MT1]}
%
\bibitem{fot:1}
Fotheringham, W.: Fotheringham's Sporting Pastimes. Anova Books. pp. 50, 2007

\bibitem{Kunkle:2}
Kunkle, D., Cooperman, G.: Twenty-Six Moves Suffice for Rubik's Cube, 
Proceedings of the International Symposium on Symbolic and 
Algebraic Computation (ISSAC '07), 2007, 
http://www.ccs.neu.edu/home/gene/papers/rubik.pdf

\bibitem{Rokicki:3}
Rokicki, T.: Twenty-Five Moves Suffice for Rubik's Cube,

http://arxiv.org/abs/0803.3435.

\bibitem{Rokicki:4}
Rokicki, T.: Twenty-Two Moves Suffice, 2008,

http://cubezzz.homelinux.org/drupal/?q=node/view/121

\bibitem{Rokicki:5}
Rokicki, T.: In search of: 21f*s and 20f*s; a four month odyssey, 2006,
http://cubezzz.homelinux.org/drupal/?q=node/view/56

\bibitem{Reid:6}
Reid, M.: Superflip requires 20 face turns, 1995,

http://www.math.rwth-aachen.de/~Martin.Schoenert/Cube-Lovers/

michael\_re\%id\_superflip\_requires\_20\_face\_turns.html

\bibitem{Reid:7}
Reid, M.: New upper bounds, 1995,

http://www.math.rwth-aachen.de/~Martin.Schoenert/Cube-Lovers/

michael\_re\%id\_new\_upper\_bounds.html

\bibitem{Radu:8}
Radu, S.: Rubik can be solved in 27f, 2006, 

http://cubezzz.homelinux.org/drupal/?q=node/view/53.

\bibitem{Flatley:9}
Flatley, J.F.: Rubik's Cube solved in twenty moves, 35 years of CPU time, Engadget, 2010.

\bibitem{Rokicki:10}
Rokicki,T., Kociemba, H., Davidson, M., Dethridge, J.: God's Number is 20, 2010, www.cube20.org.

\bibitem{Korf:11}
Korf R. E.: Macro-Operators: A Weak Method for Learning, Artificial Intelligence 26, 35-77 (1985)

\bibitem{Kautz:12}
Kautz, H.A., Selman B.: Planning as Satisfiability, European Conference on Artificial Intelligence (ECAI'92), 359--363 (1992)

\bibitem{Precosat:13}
Biere, A.: Lingeling, Plingeling, PicoSAT and PrecoSAT at SAT Race 2010,   
http://baldur.iti.uka.de/sat-race-2010/descriptions/solver\_1+2+3+6.pdf

\bibitem{chenAMO:14}
Chen, J.C.: A new SAT encoding of the at-most-one constraint, 
Proc. of the Tenth Int. Workshop of Constraint Modelling and Reformulation,
St. Andrews, Scotland, UK, 2010.

\bibitem{MoRsat:15}
Chen, J.C.: Building a Hybrid SAT Solver via Conflict-driven, 
Look-ahead and XOR Reasoning Techniques, SAT 2009, LNCS 5584, 
298-311 (2009)

\bibitem{Explorer:16}
Kociemba, H.:Cube Explorer (Windows program), 

http://kociemba.org/cube.htm

\bibitem{MPhaseSAT:17}
Chen, J.C.: The SAT solver, MPhaseSAT, submitted to SAT 2011
Competition.

\bibitem{March:18}
Heule, M., Van Maaren, H.: Effective Incorporation of double
look-ahead procedures, SAT 2007, LNCS 4501, 258--271 (2007)

\end{thebibliography}

\end{document}